\pdfoutput=1

\documentclass[11pt]{article}

\usepackage[preprint]{acl}

\usepackage{times}
\usepackage{latexsym}
\usepackage{tabularx} 

\usepackage[T1]{fontenc}

\usepackage[utf8]{inputenc}

\usepackage{microtype}

\usepackage{inconsolata}

\usepackage{graphicx}
\usepackage{multirow}
\usepackage{enumitem}
\usepackage{algorithm}
\usepackage{algorithmic}
\usepackage{amsmath}
\setitemize[1]{topsep=3pt}

\title{Quantization Meets Reasoning: Exploring LLM Low-Bit Quantization Degradation for Mathematical Reasoning}

\author{
 \textbf{Zhen Li\textsuperscript{1,4}\thanks{These authors contributed equally to this work.}},
 \textbf{Yupeng Su\textsuperscript{2}\footnotemark[1]},
 \textbf{Runming Yang\textsuperscript{3}},
 \textbf{Congkai Xie\textsuperscript{4}},
 \textbf{Zheng Wang\textsuperscript{3}},
 \textbf{Zhongwei Xie\textsuperscript{5}},
\\
  \textbf{Ngai Wong\textsuperscript{6}\protect\footnotemark[2]},
 \textbf{Hongxia Yang\textsuperscript{1,4}\protect\footnotemark[2]},
\\
 \textsuperscript{1}The Hong Kong Polytechnic University,
 \textsuperscript{2}Southern University of Science and Technology,
\\
 \textsuperscript{3}Tsinghua University,
 \textsuperscript{4}Reallm Labs,
 \textsuperscript{5}Wuhan University,
 \textsuperscript{6}The University of Hong Kong,
\\
 \textsuperscript{*}\,Corresponding authors. \quad
\texttt{nwong@eee.hku.hk, hongxia.yang@polyu.edu.hk}
\\
}

\begin{document}
\maketitle
\begin{abstract}
Large language models have achieved significant advancements in complex mathematical reasoning benchmarks, such as MATH. However, their substantial computational requirements present challenges for practical deployment. Model quantization has emerged as an effective strategy to reduce memory usage and computational costs by employing lower precision and bit-width representations. In this study, we systematically evaluate the impact of quantization on mathematical reasoning tasks. Our results demonstrate that aggressive quantization methods like AWQ and GPTQ introduce up to 32.39\% accuracy degradation (average 11.31\%) on Llama-3 models, particularly in numerical computation and reasoning planning. To address this, we introduce a multidimensional evaluation framework combining qualitative capability analysis and quantitative error assessment. We further develop targeted recovery strategies, showing that fine-tuning quantized models on only 545 task-specific examples for 3 minutes on 4 GPUs effectively restores reasoning capabilities to near full-precision levels. Additionally, our error assessment pipeline achieves 98.9\% accuracy in diagnosing and localizing errors across 3,366 failure cases, providing actionable insights for mitigating quantization-induced degradation.

\end{abstract}

\section{Introduction}
\vspace{-2mm}

Large language models (LLMs) have substantially advanced the state of mathematical reasoning in artificial intelligence, demonstrating remarkable performance on diverse tasks ranging from basic arithmetic and quantitative reasoning to intricate geometric and competition-level problems~\cite{brown2020language,chowdhery2023palm,touvron2023llama,achiam2023gpt,2023GPT4VisionSC}.
Critically, these models excel not only at producing correct final answers, but also at providing step-by-step solutions that elucidate the underlying reasoning process~\cite{lewkowycz2022solving,wei2022chain}.
Benchmarks such as MATH ~\cite{hendrycks2021measuring} highlight these capabilities, where LLMs can guide humans through complex multi-step problems with detailed reasoning chains.


However, such advancements come at a cost. The computational requirements of LLMs, both in terms of memory and latency, pose significant practical barriers ~\cite{kaplan2020scaling,hoffmann2022training,gou2024tora}. 
To address these efficiency challenges, researchers have explored model compression techniques such as pruning~\cite{lecun1989optimal,han2015deep}, knowledge distillation ~\cite{hinton2015distilling,jiao2019tinybert,yang2024llm}, and more recently, quantization ~\cite{hubara2018quantized,jacob2018quantization,yao2022zeroquant}.
Quantization reduces memory usage and computational overhead by representing weights and activations in low-bit formats (e.g., INT8), halving GPU memory consumption and nearly doubling throughput in operations like matrix multiplication and attention ~\cite{rastegari2016xnor,lin2015neural,dettmers2022gpt3}. While it performs well on standard NLP tasks with minimal performance loss ~\cite{ma2024era}, its effect on complex mathematical reasoning, requiring precise, contextually coherent, and logical steps, remains unclear, particularly for tasks like MATH or Code.


\begin{figure*}[htb]
    \centering
    \setlength{\belowcaptionskip}{-0.4cm}
    \includegraphics[width=1.0\linewidth]{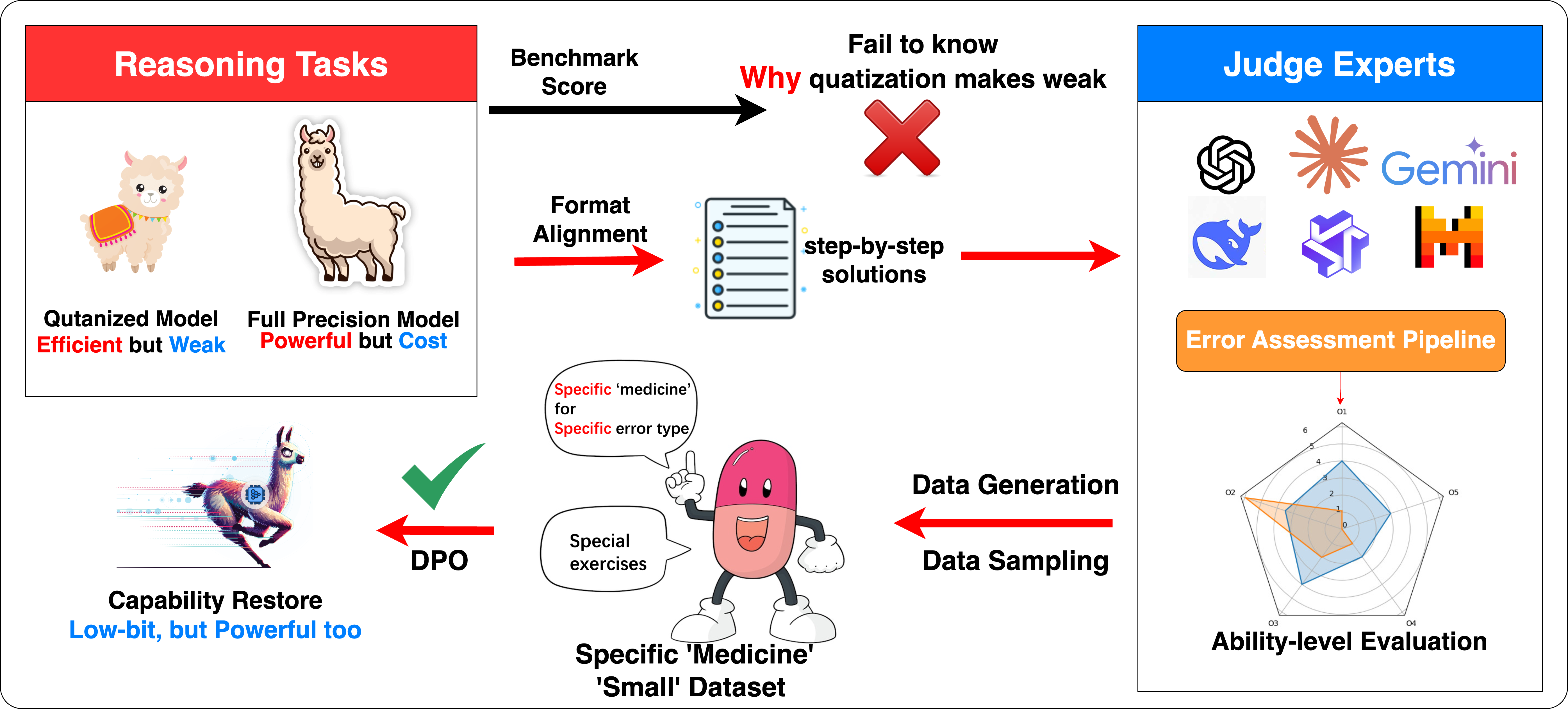}
    \caption{Pipeline of our study for evaluating and restoring reasoning capabilities in quantized models. By format alignment training and our error assessment pipeline with expert judge models, we diagnose and analyze ability-level reasoning weaknesses on model's step-by-step solutions. Based on identified error types, we sample a targeted ‘medicine’ dataset to fine-tune the model via Direct Preference Optimization (DPO), aiming to restore performance while preserving efficiency.}
    \label{fig:overview}
\end{figure*}

Prior work has hinted at potential reasoning degradation under aggressive quantization ~\cite{shen2020q,kim2021bert,Lin2023AgentSimsAO},
but a systematic understanding is lacking. Such limitations pose a stark contrast to the precision-based assumptions of advanced mathematical reasoning approaches. Models like Minerva ~\cite{lewkowycz2022solving} and reasoning strategies such as Chain-of-Thought (CoT) prompting~\cite{wei2022chain,xiong2023dq} rely heavily on high-fidelity internal representations to ensure logical consistency and correctness. 
Quantization, however, may disrupt these critical internal states. 
Meanwhile, ongoing developments in model optimization transcend simple training paradigms. Complex post-training pipelines that align models to human preferences and adapt them for specialized applications have gained traction, often involving intricate infrastructure and iterative refinement processes~\cite{schulman2017proximal,rafailov2024direct}.

Against this backdrop, the interplay between quantization, advanced inference strategies, and the underlying reasoning fidelity of LLMs emerges as a key research question. In this work, we seek to illuminate these relationships, providing insights into how quantization influences mathematical reasoning  and how we might mitigate these effects.

To summarize, in this study, we present the following key contributions:

\begin{itemize}
    \setlength{\itemsep}{-3pt}
    \item We reveal that mainstream quantization methods (AWQ, GPTQ) incur substantial reasoning degradation, with up to 32.39\% accuracy loss (average 11.31\%) on MATH for Llama-3 models, exposing critical vulnerabilities in low-bit numerical representations.
    \item  We propose a step-aligned evaluation protocol to dissect reasoning errors into four dimensions (conceptual, methodological, executional, logical), enabling granular diagnosis of quantization effects. Our automated error assessment pipeline achieves 98.9\% accuracy in categorizing 3,366 failure cases, surpassing human-in-the-loop baselines.
    \item We demonstrate that quantized models recover the performance gaps through lightweight fine-tuning on only 545 targeted examples, requiring only 3 minutes of training, thus enabling efficient deployment without sacrificing reasoning fidelity.

\end{itemize}

\vspace{-0.2cm}
\section{Related Work}

\vspace{-0.2cm}
\subsection{Quantization Techniques}
\vspace{-0.1cm}
Modern quantization methods balance efficiency and performance through minimizing the model output differences after quantization. Post-training quantization (PTQ) approaches~\cite{frantar2022gptq,lin2024awq,xiao2023smoothquant,yao2022zeroquant} enable efficient compression without retraining, while quantization-aware training (QAT) methods~\cite{hu2021lora,dettmers2024qlora} preserve task-specific performance through learnable scaling factors. Recent work extends these techniques to LLMs, though primarily evaluated on language understanding rather than reasoning tasks.

\vspace{-0.2cm}
\subsection{Mathematical Reasoning in LLMs}
\vspace{-0.1cm}
The emergence of specialized models like Minerva~\cite{lewkowycz2022solving} demonstrates LLMs' potential for advanced mathematical problem-solving. Chain-of-Thought prompting~\cite{wei2022chain} and its variants (e.g., Program-of-Thought~\cite{chowdhery2023palm}) enhance multi-step reasoning by decomposing problems into interpretable sub-steps. However, these approaches assume high-precision model representations, potentially conflicting with quantization's reduced numerical precision.

\vspace{-0.1cm}
\subsection{Model Alignment and Reasoning Optimization}
\vspace{-0.1cm}
Recent advances in model alignment integrate instruction tuning, pioneered by the FLAN framework~\cite{wei2021finetuned} and popularized through Alpaca~\cite{taori2023stanford}, with preference optimization techniques like DPO~\cite{rafailov2024direct}. Concurrently, reasoning reliability enhancements employ self-consistency voting~\cite{wang2022self} and process-based reward models~\cite{lightman2023lets}, building upon foundational work in verifiable reasoning~\cite{creswell2022faithful}. Our investigation extends these directions by analyzing how quantization affects: the instruction-following capabilities crucial for step-by-step reasoning, and the consensus-building mechanisms in ensemble-based reasoning methods.

\vspace{-0.2cm}
\section{Methodology}
\vspace{-0.2cm}
\subsection{Model Quantization}
\vspace{-0.1cm}
We conduct a comprehensive investigation into the effects of quantization techniques, examining both weight-only quantization methods (GPTQ~\cite{frantar2022gptq}, AWQ~\cite{lin2024awq}). Our evaluation encompasses various quantization configurations, specifically focusing on 4-bit weight precision with 16-bit activations (W4A16). Through the systematic application of these mainstream quantization techniques, we provide a rigorous and balanced analysis of the resulting quantized models, offering valuable insights into their performance characteristics and trade-offs.

\vspace{-0.1cm}
\subsection{Format Alignment}
\vspace{-0.1cm}
To address the challenge of inconsistent instruction following and irregular output formatting in model-generated solutions, we introduce a format alignment stage. This phase aims to instill in the model a structured, step-by-step reasoning workflow without altering its underlying mathematical knowledge. Crucially, the objective here is not to teach the model new mathematical facts, but rather to ensure strict adherence to a prescribed output format, thereby enabling reliable qualitative and quantitative analysis of reasoning capability across quantized and full-precision variants.

\begin{figure}[t]
    \centering
    \setlength{\abovecaptionskip}{-0.1cm}
    \setlength{\belowcaptionskip}{-0.3cm}
    \includegraphics[width=1.0\linewidth]{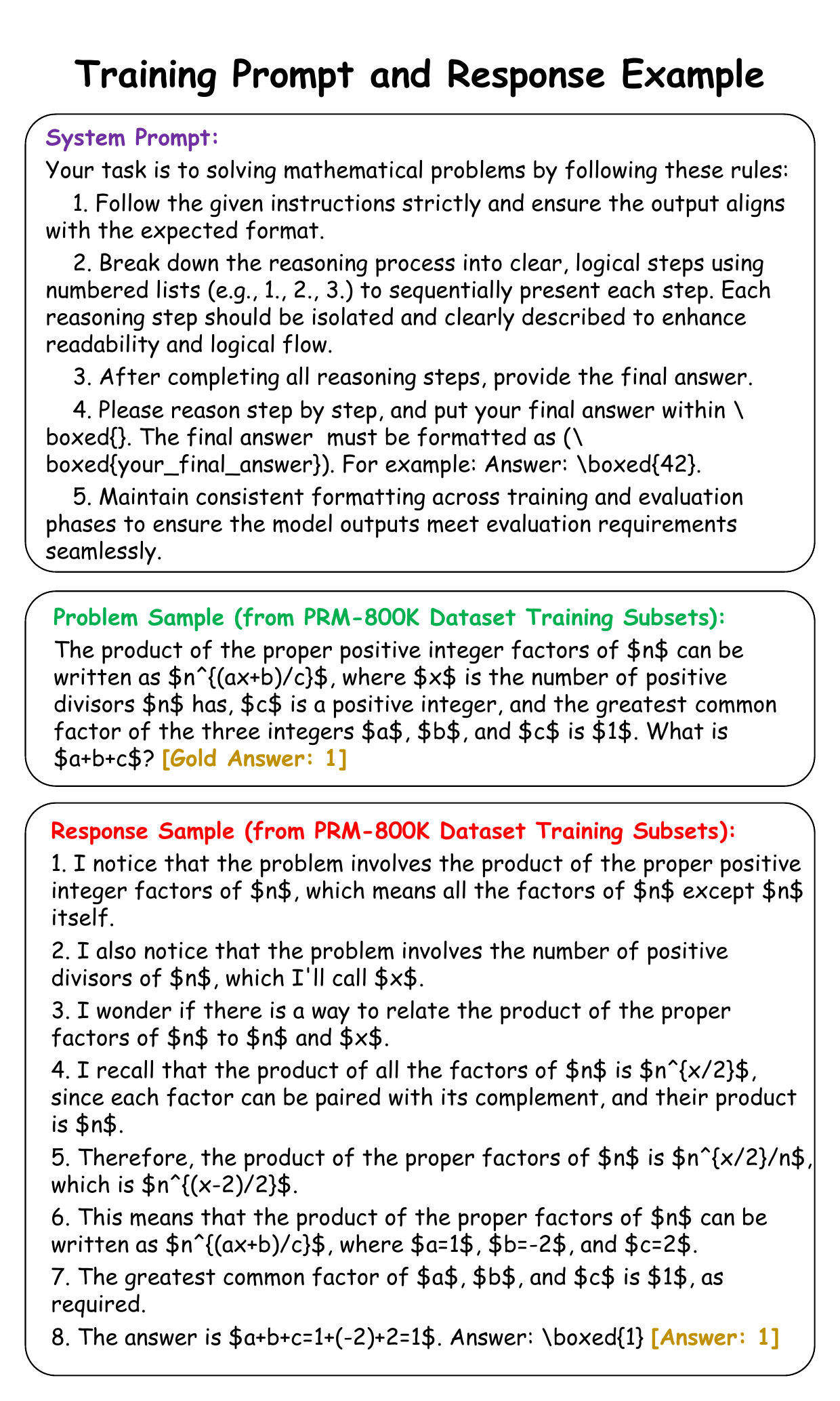}
    \caption{An example of the data used during training. The output is concatenated using added tokens, and the final answer is filled in \texttt{boxed\{\}} in the evaluation format.}
    \label{fig:prompt_example}
    \vspace{-0.4cm}
\end{figure}

We employ LoRA~\cite{hu2021lora} and QLoRA~\cite{dettmers2024qlora} for full-precision model and quantized model respectively as lightweight adaptation techniques for format alignment. By selectively updating a small subset of parameters, these methods efficiently align knowledge of step-by-step solution formats into the model’s latent space without extensive retraining. This fine-tuning enables us to observe how multi-step reasoning is preserved or altered once the model is quantized, offering deeper insights into any capability loss induced by compression.

\begin{table*}[t]
\centering
\setlength{\belowcaptionskip}{-0.5cm}
\begin{tabular*}{\linewidth}{c|cm{2.5cm}<{\centering}m{3cm}<{\centering}m{3cm}<{\centering}}
\hline
\textbf{MATH} &
  \textbf{} &
  \textbf{\begin{tabular}[c]{@{}c@{}}Vanilla\\ (BF16)\end{tabular}} &
  \textbf{\begin{tabular}[c]{@{}c@{}}AWQ\\ (W4A16)\end{tabular}} &
  \textbf{\begin{tabular}[c]{@{}c@{}}GPTQ\\ (W4A16)\end{tabular}} \\ \hline
\multirow{2}{*}{\textbf{Llama-3.1-8B-Instruct}} & \textbf{score}    & 47.2       & 41.8                     & 43.06                    \\
                                                & \textbf{$\Delta$} & \textbf{-} & $\downarrow$5.4(11.44\%) & $\downarrow$5.8(12.29\%) \\
\multirow{2}{*}{\textbf{Llama-3.2-3B-Instruct}} & \textbf{score}    & 40.2       & 38.8                     & 31.6                     \\
                                                & \textbf{$\Delta$} & \textbf{-} & $\downarrow$2.16(4.97\%) & $\downarrow$4.0(10.42\%) \\
\multirow{2}{*}{\textbf{Llama-3.2-1B-Instruct}} & \textbf{score}    & 18.6       & 14.26                    & 14.94                    \\
                                                & \textbf{$\Delta$} & \textbf{-} & $\downarrow$2.8(16.67\%) & $\downarrow$6.8(32.39\%) \\ \hline
\end{tabular*}
\caption{Evaluation results of full-precision models and 4-bit weight, 16-bit activation quantization methods (AWQ and GPTQ) on the MATH benchmark. All three model scales exhibit varying degrees of performance degradation, posing challenges in solving mathematical reasoning problems. Notably, smaller-scale models suffer a performance drop exceeding 30\%.}
\label{MATH_result}
\end{table*}

For training, we utilize the PRM800K dataset~\cite{lightman2023let}, which provides 800K step-level correctness annotations from 75K solutions to 12K problems. These annotations supply granular, step-by-step reasoning trajectories, equipping models to separate complex problem-solving processes into well-defined stages. To reinforce this structure, we adopt a consistent system prompt across training and evaluation, ensuring that the boundaries of logical steps and final answers are clearly delineated. This consistent, step-by-step alignment is a necessary foundation for our subsequent qualitative and quantitive analyses of quantization-induced degradation in mathematical reasoning.

\vspace{-0.1cm}
\subsection{Analyses for Reasoning Errors}
\vspace{-0.1cm}
To investigate the underlying reasons for the performance degradation observed in the quantized models, we categorized the errors into seven specific error types in four major error categories.:
\begin{itemize}[leftmargin=1em]
\setlength{\parskip}{-1pt}
    \item \textbf{Conceptual Errors}
    \begin{itemize}
            \item \textbf{Conceptual misunderstanding}: Inadequate grasp of fundamental concepts or principles, leading to an incorrect framing of the problem or misguided approach to the solution.
            \item \textbf{Contextual oversight}: Failure to account for relevant contextual factors or domain constraints (e.g., physical limits, geometric considerations) that significantly affect the problem’s outcome.
    \end{itemize}
    \item \textbf{Method Errors}
    \begin{itemize}
            \item \textbf{Procedural error}: Errors in executing or adhering to prescribed procedures (e.g., failure to follow a standard algorithm or overlooking a required step), resulting in incomplete or invalid solutions.
            \item \textbf{Formula rule error}: Improper application of mathematical rules, theorems, or formulae (including misapplication of a formula to an inappropriate scenario), which undermines the validity of the derived results.
    \end{itemize}
    \item \textbf{Execution Errors}
    \begin{itemize}
            \item \textbf{Computational error}: Mistakes in arithmetic or algebraic manipulation, such as incorrect summation, factorization, or expansion, that compromise the accuracy of the final answer.
            \item \textbf{Symbolic manipulation error}: Inaccurate handling of symbols or expressions, including mislabeling variables or misinterpreting symbolic transformations, leading to an incorrect representation of the problem.
    \end{itemize}
    \item \textbf{Reasoning Errors}
    \begin{itemize}
            \item \textbf{Logical reasoning error}: Breakdown in the logical chain of thought, where the inference steps do not follow coherently or omit essential logical links, causing a disconnect between premises and conclusion.
    \end{itemize}
\end{itemize}

\vspace{-0.1cm}
\subsection{Capability Restore}
\vspace{-0.1cm}
Building upon our systematic error analysis, we develop targeted capability restoration strategies to address quantization-induced performance degradation. For each identified error category (conceptual, methodological, executional, and logical),  we create focused training subsets by pairing failed step-wise responses with correct responses to form a dataset of positive and negative pairs:

\begin{itemize}[leftmargin=1em]
\setlength{\parskip}{-2pt}
\item \textbf{Conceptual Restoration}: High-context problems requiring domain-specific constraints.
\item \textbf{Methodological Restoration}: Multi-step derivations with annotated procedure adherence.
\item \textbf{Executional Restoration}: Computation-intensive problems with stepwise verification.
\item \textbf{Logical Restoration}: Chain-of-thought reasoning with explicit inference links.
\end{itemize}

\begin{figure}[t]
    \setlength{\abovecaptionskip}{-0.1cm}
    \setlength{\belowcaptionskip}{-0.1cm}
    \centering
    \includegraphics[width=\linewidth]{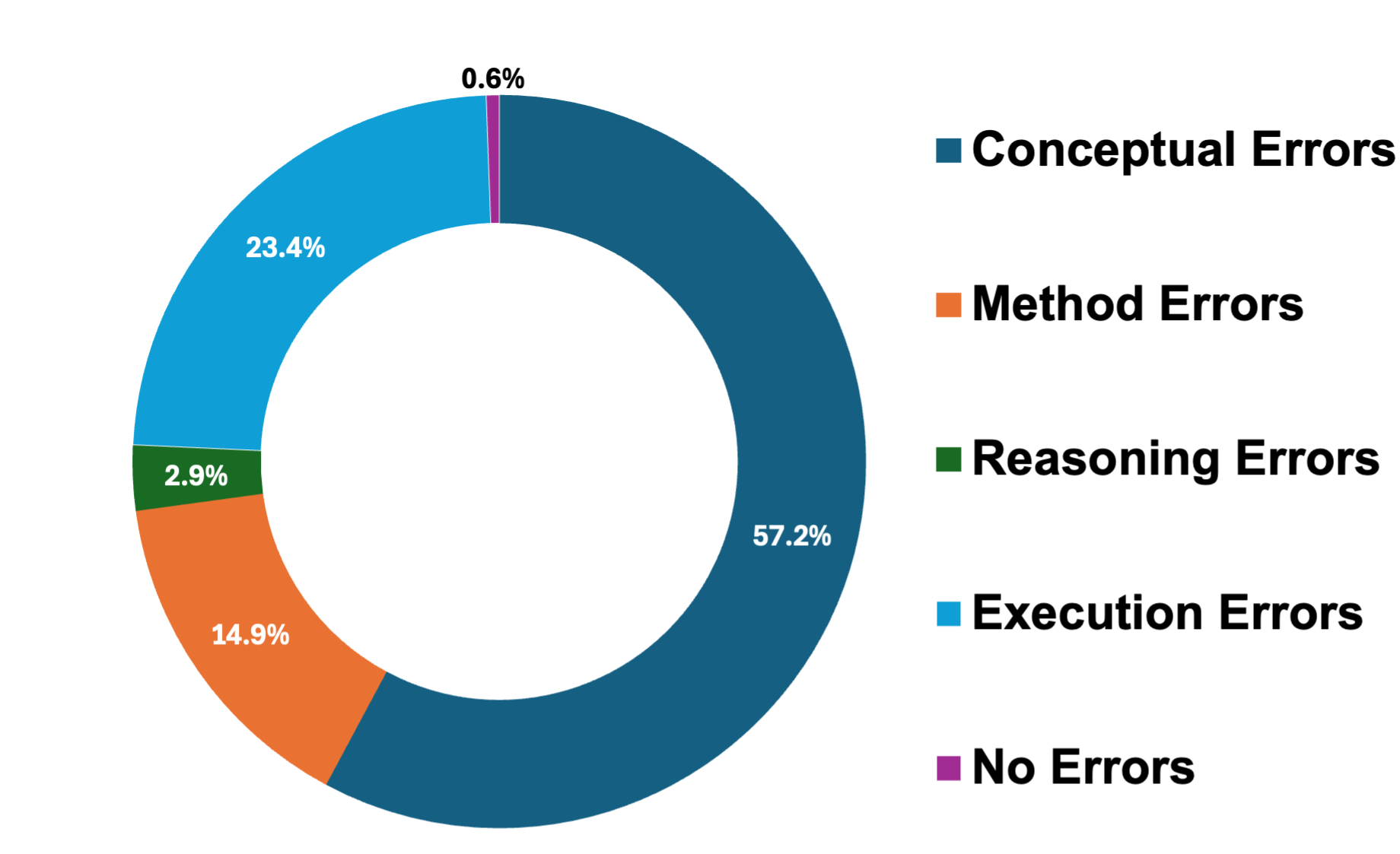}
    \caption{Distribution of error types in quantized models, highlights the dominant error types affecting mathematical reasoning in quantized models.}
    \label{fig:pieFig}
\end{figure}

\begin{figure}[htbp]
    \setlength{\belowcaptionskip}{-0.3cm}
    \centering
    \includegraphics[width=\linewidth]{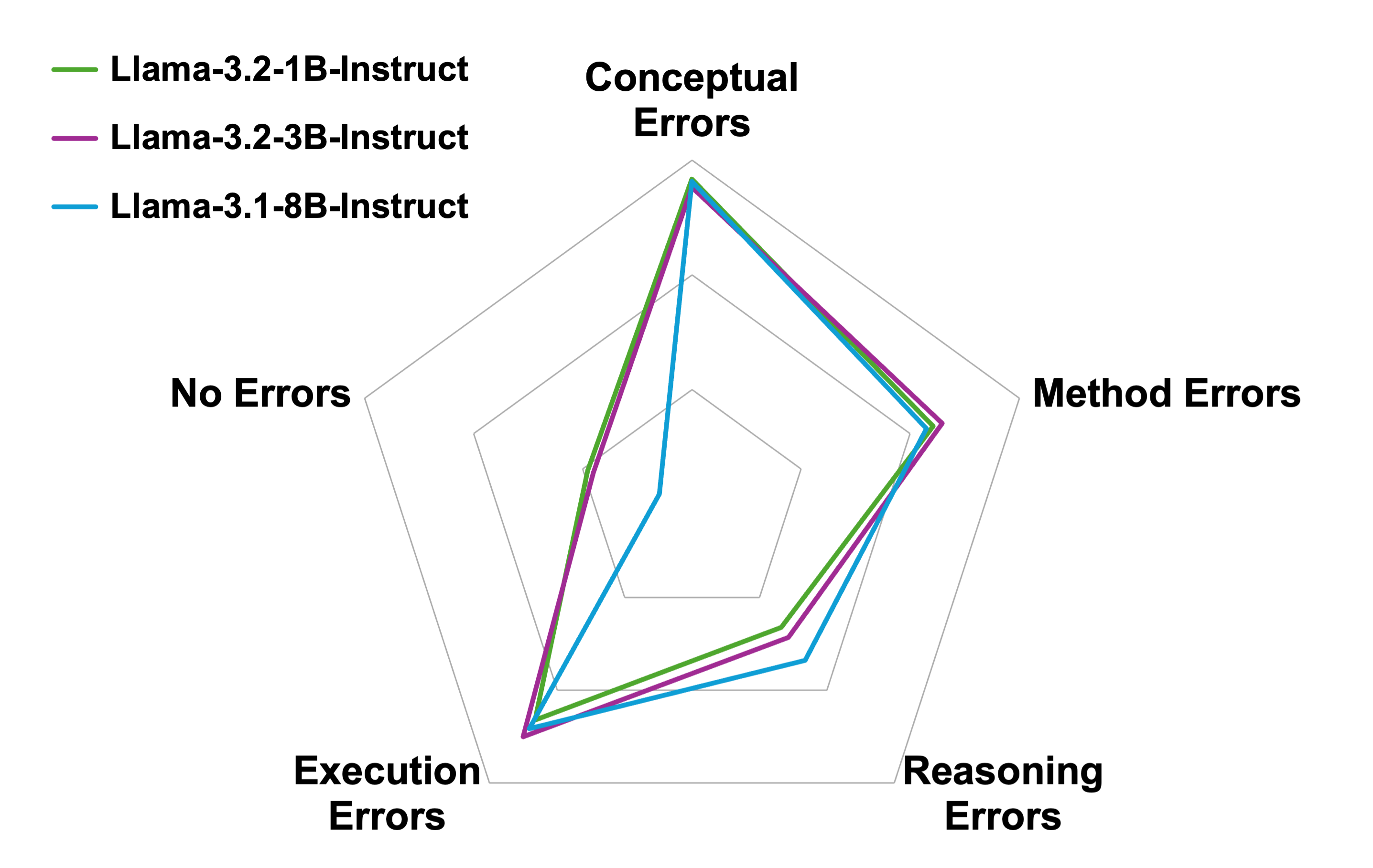}
    \caption{The radar plot of error distributions across different scale Llama models on the MATH benchmark.}
    \label{fig:radarFig}
\end{figure}

We implement Direct Preference Optimization (DPO) with parameter-efficient adaptation, using QLoRA for quantized models. This dual approach enables simultaneous optimization of two objectives: (1) maximizing the likelihood of correct solution trajectories through supervised fine-tuning on error-specific subsets, and (2) addressing the model’s capability gaps while leveraging the undiminished foundational abilities of the quantized model to complete the solution, thereby enhancing overall mathematical reasoning performance.

Our recovery process comprises three phases:
\vspace{-0.2cm}
\begin{enumerate}[leftmargin=1.5em]
\setlength{\itemsep}{-3pt}
	\item \textbf{Data Extraction and Construction:} We extract and construct a dataset from the quantized models’ failed responses across different model scales.
	\item \textbf{Data Deduplication and Cross-Scale Sampling:} To avoid redundancy, we deduplicate the dataset and perform cross-scale sampling based on error rates. This facilitates cross-error joint training with a mix of problem difficulties.
	\item \textbf{DPO Training on the Aligned Model:} Building on the format-aligned model, we further train using a DPO approach.
\end{enumerate}
This phased approach achieves significant capability recovery while introducing less than 1\% additional parameter updates relative to the original model sizes. Our experiments demonstrate that targeted restoration using as few as 545 error-specific examples per category can bridge the performance gap between quantized and full-precision models. Remarkably, even training with a dataset comprising only 323 samples for a single error type can restore up to 9\% of the model’s capability, confirming that localized capability losses can be efficiently remediated through structured retraining.




\begin{algorithm}[H]
\caption {Pseudocode of our approach.}

\textbf{Input:} Full-precision model $M$, Quantization method $Q$, Step-annotated dataset $D_{train}$, Evaluation dataset $D_{test}$, Expert models $\{E_1...E_n\}$

\textbf{Output:} Enhanced quantized model $M_{enhanced}$

\begin{algorithmic}[1]

\STATE \textbf{Step 1: Model Quantization}
\STATE Apply quantization method $Q$ to model $M$:
\STATE $M_Q \gets Q(M)$ 

\STATE \textbf{Step 2: Format Alignment}
\STATE Fine-tune $M_Q$ with QLoRA on $D_{train}$:
\STATE $M_A \gets \text{QLoRA}(M_Q, D_{train})$ 

\STATE \textbf{Step 3: Error Evaluation}
\STATE Generate outputs $O$ on $D_{test}$:
\STATE $O \gets M_A(D_{test})$
\STATE Initialize error set $E \gets \emptyset$

\FOR{each output $o_i \in O$}
    \STATE \textbf{Multi-Expert Evaluation:}
    \STATE Collect evaluations $\{E_1(o_i)...E_n(o_i)\}$
    \IF{Majority vote $\neq$ DeepSeek-R1 \textbf{and} votes $<$ 4}
        \STATE Flag $o_i$ as $\text{Rejected}$
        \STATE $E \gets E \cup \{\text{ErrorType}(o_i)\}$ 
    \ENDIF
\ENDFOR

\STATE \textbf{Step 4: Human Verification}
\STATE Random sample $s \gets 2\%$ of $E$:
\STATE $E_{sample} \gets \text{RandomSample}(E, s)$
\STATE $E_{verified} \gets \text{HumanVerify}(E_{sample})$

\STATE \textbf{Step 5: Capability Restoration}
\FOR{each error type $t \in E_{verified}$}
    \STATE Generate targeted dataset $D_t$:
    \STATE $D_t \gets \text{CurateData}(t, D_{train})$ 
    \STATE Apply DPO training:
    \STATE $M_A \gets \text{DPO}(M_A, D_t)$
\ENDFOR

\STATE \textbf{Return:} $M_{enhanced} \gets M_A$
\end{algorithmic}
\end{algorithm}

\vspace{-0.2cm}
\section{Experiments}
\subsection{General Setup}
All experiments were conducted on a cluster comprising 4 Nvidia 80G A800 GPUs.

\noindent\textbf{Base Model \& Scale}


Given the general applicability of our approach to mainstream models and the versatility across different model sizes, we select three distinct models from the Llama 3 series as the foundation for our experiments. 
Specifically, we choose models of varying sizes, namely 8B for Llama-3.1 and 1B/3B for Llama-3.2, all Instruct version. 

For all of these models, we observe significant math accuracy degradation resulting from quantization, and we also demonstrate the powerful effect of our proposed fast reconstruction method.

\noindent\textbf{Quantization Strategy.} Our experiments employ two mainstream quantization techniques: AWQ and GPTQ~\cite{frantar2022gptq,lin2024awq}, both known for their strong performance in model compression. 
These methods have led to numerous low-precision models widely adopted in practice. Our work aims to provide complementary insights for community.

\noindent\textbf{Training Strategy.} We use a sequence length of 1024 and in both the format alignment and capability restoration stages, we employed a global batch size of 16 and a warm-up period spanning 3\%. All models were trained for 3 epochs.

\noindent\textbf{Evaluation Strategy.} We evaluated our models using the OpenCompass evaluation framework~\cite{2023opencompass}, maintaining consistency in both the global batch size and the prompt with those used during training. This setup ensures a fair comparison across all models.

\subsection{Format Alignment}

\textbf{Setting.} We applied LoRA fine-tuning to the vanilla model, while QLoRA~\cite{dettmers2024qlora} was employed for the quantized models. The LoRA rank was set to 32 for the format alignment stage. During training, the system prompt strictly enforced a step-wise output of the reasoning process, and outputs that deviated from the required format were penalized.

\begin{figure}[htbp]
    \setlength{\abovecaptionskip}{-0.1cm}
    \setlength{\belowcaptionskip}{-0.3cm}
    \includegraphics[width=1.0\linewidth]{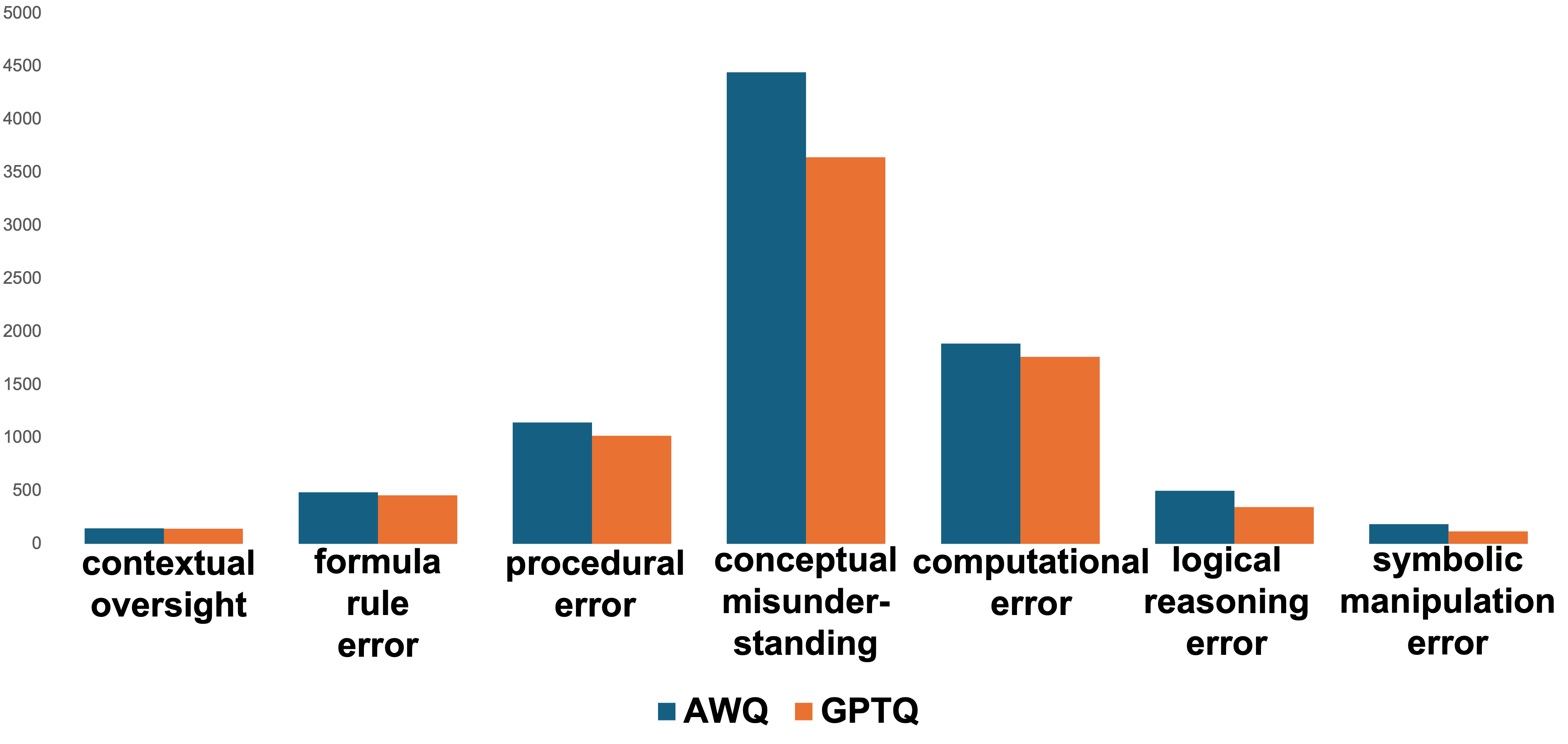}
    \caption{Comparative analysis of error types for quantized models (AWQ-W4A16 and GPTQ-W4A16) across three model scales.}
    \label{fig:columnFig}
\end{figure}

\subsection{Analyses for Reasoning Errors}

\textbf{Error Assessment Pipeline.} To facilitate a rigorous and scalable evaluation of quantization-induced errors in reasoning tasks, we developed an automated assessment pipeline powered by state-of-the-art language models. This pipeline systematically processes model outputs and classifies errors according to our predefined \textit{error\_types\_list} taxonomy. By leveraging a pre-trained transformer as the core evaluator, we reduce subjective bias and ensure consistent, reproducible error analyses across all experimental conditions. Furthermore, the computational scoring framework supports high-throughput performance assessment while preserving granularity in error categorization.

Our evaluation pipeline comprises three primary stages:
\begin{enumerate}
    \item \textbf{Model Judge:} For each instance in which a quantized model produces an incorrect answer, we employ a dedicated “judge model” to analyze the error. The judge model is tasked with: (a) identifying the first occurrence of an error, (b) specifying the exact step where the error is introduced, (c) assigning an error category based on a nested classification scheme, and (d) providing an explanation along with a confidence score for its determination.
    \item \textbf{Majority Vote:} To mitigate hallucinations and enhance evaluation stability, we implement a majority voting mechanism across five models: DeepSeek-R1, GPT-4o, GPT-4, Qwen-Max, and DeepSeek-V3. DeepSeek-R1’s assessment is treated as the baseline, and its results are compared with the majority vote from the other models. Instances of disagreement are flagged for further review, ensuring consistency and minimizing spurious judgments.
    \item \textbf{Human Verification:} For cases with conflicting assessments from the majority vote, manual review is conducted. Additionally, we randomly sample 2\% of the evaluated cases to verify the accuracy and consistency of the automated judgments.
\end{enumerate}

This pipeline generally aligns with human error analysis and minimizes misclassifications and inconsistencies. 
However, we also encountered unexpected scenarios. For instance, when the canonical answer is \verb|"\\frac{11}{2}"| but the quantized model outputs \verb|"5.5"| the judge model sometimes erroneously concludes that there is "No Error" due to subtle discrepancies in reasoning or formatting even if we do not give this type of judgment. Such findings underscore both the robustness of our judge framework.


\begin{table*}[ht]
\centering
\begin{tabular*}{\linewidth}{c|ccm{2.5cm}<{\centering}m{2.5cm}<{\centering}m{2.5cm}<{\centering}}
\hline
\textbf{MATH-500} &  & \textbf{} & \textbf{\begin{tabular}[c]{@{}c@{}}Vanilla\\ (BF16)\end{tabular}} & \textbf{\begin{tabular}[c]{@{}c@{}}AWQ\\ (W4A16)\end{tabular}} & \textbf{\begin{tabular}[c]{@{}c@{}}GPTQ\\ (W4A16)\end{tabular}} \\ \hline
\multirow{2}{*}{\textbf{Llama-3.1-8B-Instruct}} & \multirow{2}{*}{\textbf{Aligned}} & \textbf{score} & 37.36 & 30.4 & 30.8 \\
 &  & \textbf{$\Delta$} & - & $\downarrow$6.94(18.58\%) & $\downarrow$6.56(17.56\%) \\
\multirow{2}{*}{\textbf{Llama-3.2-3B-Instruct}} & \multirow{2}{*}{\textbf{Aligned}} & \textbf{score} & 37.4 & 34.4 & 31.4 \\
 &  & \textbf{$\Delta$} & - & $\downarrow$3.0(8.02\%) & $\downarrow$6.0(16.04\%) \\
\multirow{2}{*}{\textbf{Llama-3.2-1B-Instruct}} & \multirow{2}{*}{\textbf{Aligned}} & \textbf{score} & 21 & 14.2 & 14.94 \\
 &  & \textbf{$\Delta$} & - & $\downarrow$6.8(32.39\%) & $\downarrow$6.2(29.52\%) \\ \hline
\end{tabular*}

\caption{Evaluation results of format-aligned full-precision models and quantization methods (AWQ and GPTQ) on the MATH-500 benchmark. After alignment, the model is required to output responses strictly following the numbered step-wise format, resulting in slight changes in evaluation score.}
\end{table*}

\noindent\textbf{Quantitative analyses.} Table \ref{MATH_result} presents a performance comparison between full-precision and quantized models on the MATH dataset across three model scales. We report both the performance (Score) and the corresponding accuracy degradation ($\Delta$) induced by quantization. The vanilla model serves as the upper baseline, while quantized models exhibit varying degrees of performance loss. Notably, our results indicate that the 1B model is particularly sensitive to quantization.

Based on the set of questions answered incorrectly by the quantized models, we conducted further quantitative analyses. Figure \ref{fig:pieFig} and \ref{fig:columnFig} illustrates the overall distribution of the four error categories. Conceptual errors account for the majority, representing more than half of all observed mistakes. Execution errors follow next. Our case study analyses reveal that quantized models are particularly deficient in basic arithmetic operations, with multiplication errors showing the most significant performance decline. We hypothesize that these computation errors, especially in models quantized to 4-bit weights, are likely attributable to low-bit precision issues such as overflow and underflow, which propagate inaccuracies throughout multi-step calculations. Additional illustrative cases are provided in the Appendix.

Intuitively, one might expect that reasoning errors would be more frequent given the inherent complexity of mathematical reasoning. However, our analysis shows that reasoning errors are less common. This observation may reflect a “survivor effect,” whereby earlier errors, such as logical or computational failures, prevent the model from reaching subsequent steps where boundary conditions would be tested. For precise statistics, we recorded only the first error type encountered in each instance. These findings suggest that quantization disrupts the step-by-step reasoning process, thereby exposing the models’ limitations in managing complex mathematical tasks. The underlying causes likely stem from limited numerical precision and the accumulation of quantization noise in low-bit formats, which impair the representation of long-term dependencies and intermediate values during complex reasoning.

We hypothesize that small losses in specific capability dimensions can be compensated by fine-tuning on a minimal subset of targeted data. This approach could potentially restore overall performance by addressing the specific weaknesses introduced by quantization. For instance, targeted training on multi-step reasoning tasks could reduce step omissions by reinforcing the model's ability to retain intermediate results. By strategically addressing these specific failure modes, it may be possible to recover and enhance the overall capabilities of quantized models, leveraging their efficiency while minimizing performance degradation.

\setlength{\tabcolsep}{4.8pt}
\begin{table*}[t]
\centering
\setlength{\belowcaptionskip}{-0.5cm}
\begin{tabular}{l|ccccccc}
\hline
 & \multicolumn{2}{l}{\textbf{Llama-3.2-1B-Instruct}} & \multicolumn{2}{l}{\textbf{Llama-3.2-3B-Instruct}} \\
\textbf{Full Precision} & \multicolumn{2}{c}{18.6} & \multicolumn{2}{c}{40.2} \\ \hline
\textbf{} & \textbf{AWQ} & \textbf{GPTQ} & \textbf{AWQ} & \textbf{GPTQ} \\ \cline{2-5} 
\textbf{Aligned} & \textbf{14.26} & \textbf{14.94} & \textbf{38.8} & \textbf{31.6} \\
\textbf{Setting 0} & 17.2 & 18.6 & 41.6 & 38.2 \\
\textbf{Setting 1} & 15.8 & 18.0 & \textbf{{\color[HTML]{FE0000} 41.2}} & 36.2  \\
\textbf{Setting 2} & \textbf{{\color[HTML]{FE0000}16.6}} & 16.6 & 37.2 & 34.4  \\
\textbf{Setting 3} & 15.2 & 15.4 & 35.2 & 33.0  \\ \hline
\end{tabular}
\label{table: ablation_study_setting}
\caption{Ablation study settings and evaluation results on the MATH-500 dataset. Setting 0: All error cases; Setting 1: Conceptual error cases only; Setting 2: Methodological error cases only; Setting 3: Execution error cases only.}
\end{table*}

\vspace{-0.1cm}
\subsection{Capability Restore}
\vspace{-0.1cm}

\textbf{Dataset Sampling.} 
We constructed the dataset for capability restoration by sampling from all MATH problems that were answered correctly by the full-precision model but incorrectly by the quantized model. Data sampling was guided by proportional representation and consistency with expert model major vote outcomes. To enhance data validity, we performed deduplication and prioritized cases where the major vote results were most consistent and the errors were more pronounced. We aimed to maintain a compact dataset to reduce training resource consumption. Ultimately, we effectively sampled 545 examples from 3329 quantized model failure cases. The full-precision model’s correct answers were used as positive examples, while the quantized model’s incorrect answers served as negative examples, forming our DPO dataset.

\noindent \textbf{Setting.} 
Training for capability restoration stage was conducted on the format-aligned model. For this stage, the LoRA rank was still set to 32, and a cosine learning rate schedule was employed with a warm-up ratio of 0.1. We used a sigmoid function as the preference loss function, with a global batch size of 8 and a learning rate of $1 \times 10^{-6}$. For evaluation, to ensure fairness and mitigate data leakage risks, we used the MATH-500 dataset. MATH-500 comprises 500 problems that are not included in the PRM800K dataset and were randomly sampled from the MATH dataset, thereby ensuring a similar difficulty distribution.

\begin{table}[htbp]
\setlength{\belowcaptionskip}{-0.1cm}
\resizebox{\linewidth}{!}{
\begin{tabular}{lcccc}
\hline
 & \multicolumn{2}{l}{\textbf{Llama-3.2-1B-Instruct}} & \multicolumn{2}{l}{\textbf{Llama-3.2-3B-Instruct}} \\ \hline
\textbf{Full Precision} & \multicolumn{2}{c}{\textbf{21}} & \multicolumn{2}{c}{\textbf{37.4}} \\
\textbf{} & \textbf{AWQ} & \textbf{GPTQ} & \textbf{AWQ} & \textbf{GPTQ} \\ \hline
\textbf{Format Aligned} & 14.26 & 14.94 & 38.8 & 31.6 \\
\textbf{Restored} & 18.2 & 18.8 & 41.6 & 38.2 \\ \hline
\end{tabular}}
\caption{Capability restoration experiment results on the MATH-500 benchmark. Using only 545 samples within a few minutes.}
\label{table:restore_result}
\end{table}

\noindent\textbf{Results.} We conducted rapid capability restoration training on two smaller-scale models. Table \ref{table:restore_result}  presents the restoration results using a dataset of 545 sampled examples. Due to the limited data volume, training required only 2–4 minutes on 4 GPUs, demonstrating minimal resource consumption. Notably, the quantized models exhibited marked improvements on this dataset.

\vspace{-0.1cm}
\subsection{Ablation Study}
\vspace{-0.1cm}
We conducted quantitative analyses focusing on the earliest error type observed during each model’s reasoning process. We hypothesize that quantized models may exhibit a “bucket effect” in reasoning tasks, wherein they still perform well on instruction following yet fail to solve complex reasoning problems. To investigate this, we isolated the three most frequently occurring early error types for ablation experiments: (1) Conceptual error cases only, (2) Method error cases only, (3) Execution errors only. (4) All error types cases. This targeted analysis allowed us to assess the specific contributions of each error category to the overall degradation in reasoning performance, thereby providing insights into how quantization selectively impacts different aspects of model capability.

The ablation experiments results reveal that even when training on datasets containing only a single type of error, the model’s capabilities can be significantly improved. For instance, using only data associated with conceptual errors enabled the quantized model to recover reasoning performance comparable to that of the full-precision model. Notably, this subset comprised merely 323 examples and takes very few minutes for training, which required minimal computational resources. These findings lend support to our hypothesis: quantization does not diminish the model’s fundamental abilities but selectively degrades specific dimensions that are crucial for solving reasoning tasks, such as those encountered in mathematical problem solving.

\vspace{-0.2cm}
\section{Conclusion}
\vspace{-0.2cm}
This study investigates the performance degradation of low-bit quantized LLMs in complex mathematical reasoning tasks. We find that quantization disproportionately impacts conceptual and computational reasoning, primarily due to precision loss and noise propagation in low-bit representations. To address these issues, we introduce a multidimensional evaluation pipeline and demonstrate that targeted capability restoration strategies can significantly recover performance with minimal training using our specific 'medicine' datasets. Our parameter-efficient adaptation methods, such as QLoRA and DPO, offer practical solutions for resource-constrained deployments. These findings highlight the trade-off between efficiency and reasoning fidelity, providing actionable insights for optimizing quantized LLMs. Future work will explore extending this framework to other reasoning domains and investigating potential advantages of reduced precision in specific scenarios.
\section{Limitations \& Future Works}
In this study, we investigated the performance degradation of quantized models on the competitive mathematical reasoning benchmark MATH through both qualitative and quantitative analyses, employing seven distinct capability dimensions for evaluation. However, our work is constrained by limited time and computational resources, which restrict the scale of our experiments and prevent comprehensive case studies or ablation analyses across models of varying sizes. Additionally, we did not explore whether the planning failures and computational errors observed in low-bit models are caused by storage overflows resulting from low-precision data formats through knowledge exposure experiments.

Moreover, we identified intriguing phenomena where quantized models outperform full-precision models in certain specific mathematical solving steps. However, we did not conduct extensive experiments to further investigate these enhancements. As a short-term objective, we plan to perform more targeted analyses on the degraded capabilities of quantized models and explore efficient methods for their recovery, incorporating more human involvement in case studies to uncover deeper insights. Furthermore, we aim to extend our evaluation to other types of reasoning tasks, assessing quantized models from the perspective of their inherent properties to achieve a more comprehensive understanding of their performance across diverse reasoning scenarios.

\bibliography{custom}

\appendix

\newpage

\clearpage
\section{Solutions by Quantization Models}
\subsection{Execution Errors: Computational error}

In Case \#93, the task was to find the value of $c$ for which the circle defined by $x^2-10x+y^2+6y+c=0$ has a radius of $1$. The correct approach involves completing the square, resulting in $(x-5)^2+(y+3)^2=-c+34$, and solving $-c+34=1$ to find $c=33$ . However, the GPTQModel made a computational error by incorrectly rearranging the equation as $34-c=1$ leading to the wrong answer $c=35$. This error stemmed from mishandling the sign of $c$ during algebraic manipulation, highlighting the importance of careful sign management in equation solving. The case \#93 shows the \textbf{Computational Error}.

\begin{figure}[h]
\setlength{\abovecaptionskip}{-0.1cm}
\setlength{\belowcaptionskip}{-0.1cm}
    \centering
    \includegraphics[width=\linewidth]{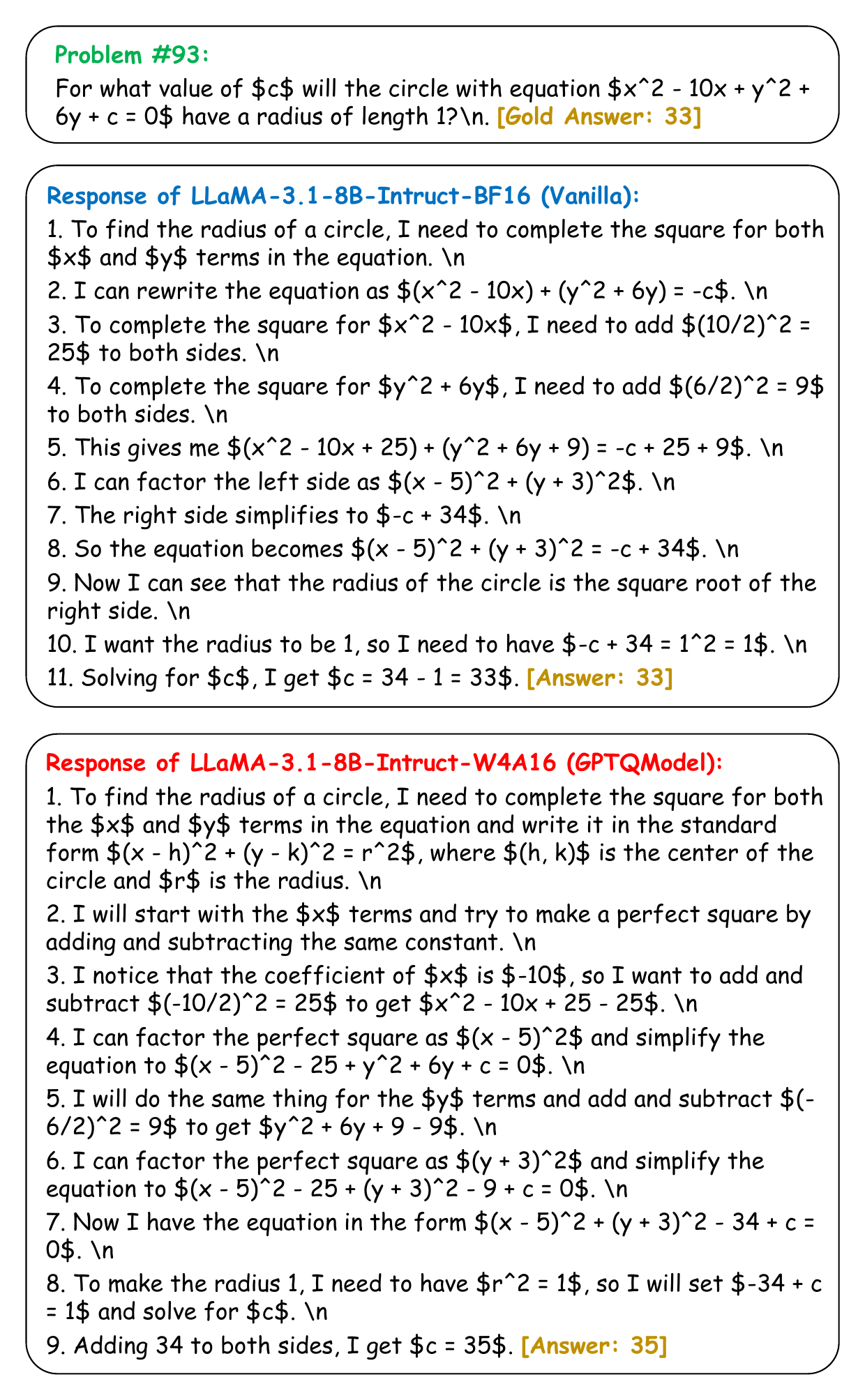}
    \caption{Problem \#93: Computational Error.}
    \label{fig:computational_error}
\end{figure}

\newpage
\subsection{Reasoning Errors: Logical reasoning error}

In Problem \#128, the GPTQModel made a logical reasoning error when calculating the minimum number of miles Suzanne could walk in February. While the model correctly identified that February has 28 or 29 days, it missed the 27th day when calculating her walking schedule every third day, leading to an incorrect conclusion of 8 walking days and 32 miles. In reality, the correct number of walking days is 9 (3, 6, 9, 12, 15, 18, 21, 24, 27), resulting in a total of $9 \times4=36$ miles. This error highlights the model's logical reasoning gap in iterating through sequential intervals accurately. The case \#128 shows the \textbf{Logical Reasoning Error}.
\begin{figure}[h]
\setlength{\abovecaptionskip}{-0.1cm}
\setlength{\belowcaptionskip}{-0.1cm}
    \centering
    \includegraphics[width=\linewidth]{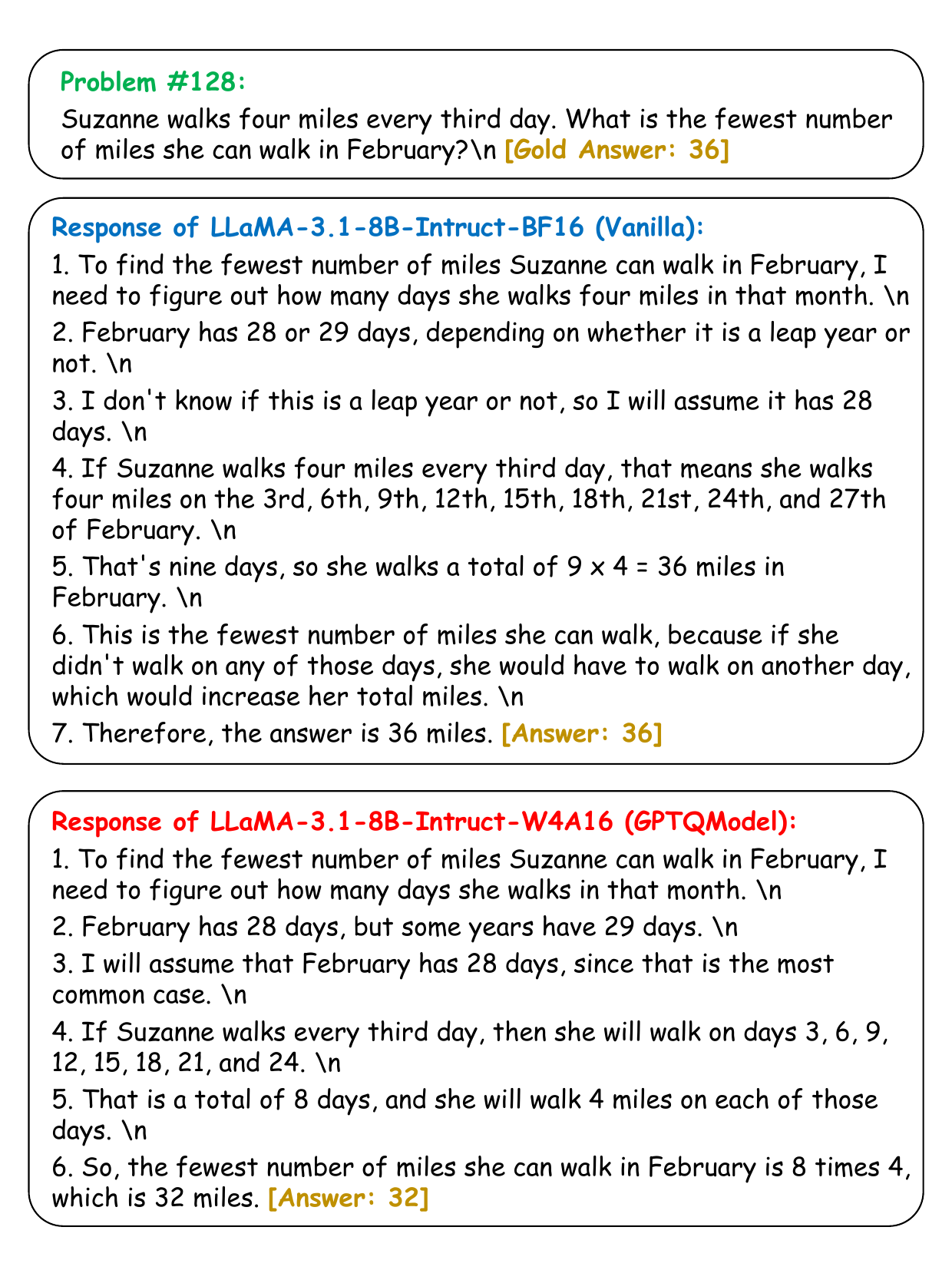}
    \caption{Problem \#128: Logical Reasoning Error.}
    \label{fig:logical_reasoning_error}
\end{figure}

\newpage
\section{Cross Validation}
\subsection{Conflict Judges}

The case \#342 shows how we check the Error type of conflicting cases with the help of Judge models and human for double-check.

We use five models cross-validation to determine the type of model reasoning errors after quantization. Specifically, we use Deepseek-R1, Deepseek-V3, ChatGPT, GPT-4o and Qwen-Max as judge models, and use five models to determine the cause of the error in the reasoning process, and use the majority voting decision method to finally determine the error type.

\vspace{-0.3cm}
\begin{figure}[h]
\setlength{\abovecaptionskip}{-0.1cm}
\setlength{\belowcaptionskip}{-0.5cm}
    \centering
    \includegraphics[width=\linewidth]{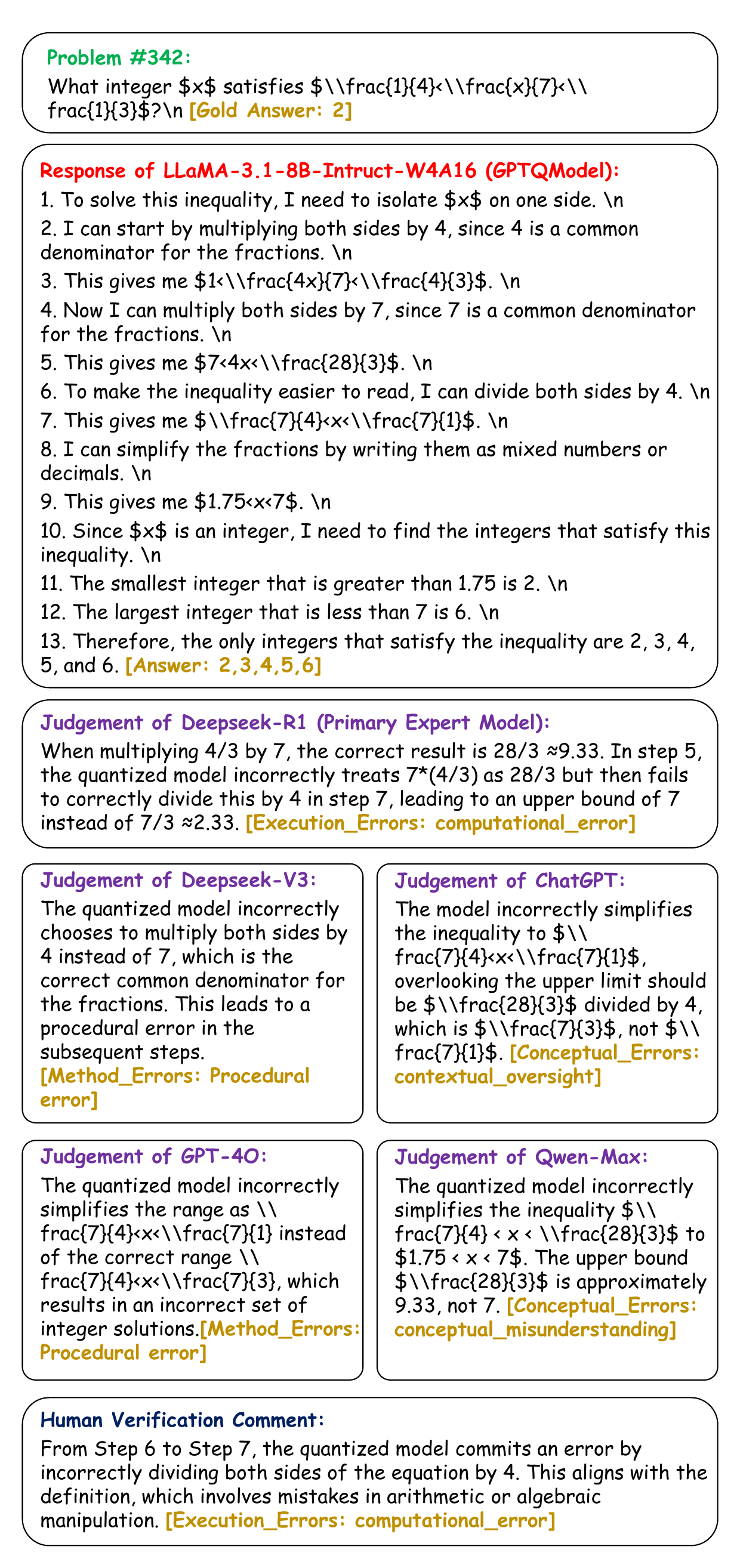}
    \caption{Problem \#342: Conflict Judges.}
    \label{fig:conflict_judges}
\end{figure}

\newpage
At the same time, we use manual intervention verification to randomly select about 2\% of conflicting cases, manually determine the error type and compare it with the judgment of the judge model to ensure that the judge's vote is at a relatively correct level.

\subsection{No Errors}

Although we select the extraction scripts carefully, there are still few samples where the model's output results are difficult to be correctly identified, resulting in special "No Errors" situations. 
This situation can usually be identified by the excellent Judge model, such as case \#3812 above. The Deepseek-R1 model can reply \textbf{No Errors} after careful comparison, and we will also correct these cases among results.

Examples of easy confusion often appear in the choice of representation of decimals and fractions, including the expression between natural syntax and programming syntax, 
e.g., 1/2 vs 0.5  vs \texttt{\textbackslash frac\{1\}\{2\}} vs 5E-01 vs \verb|5 x 10^-1|.

\begin{figure}[h]
\setlength{\abovecaptionskip}{-0.1cm}
\setlength{\belowcaptionskip}{-0.1cm}
    \centering
    \includegraphics[width=\linewidth]{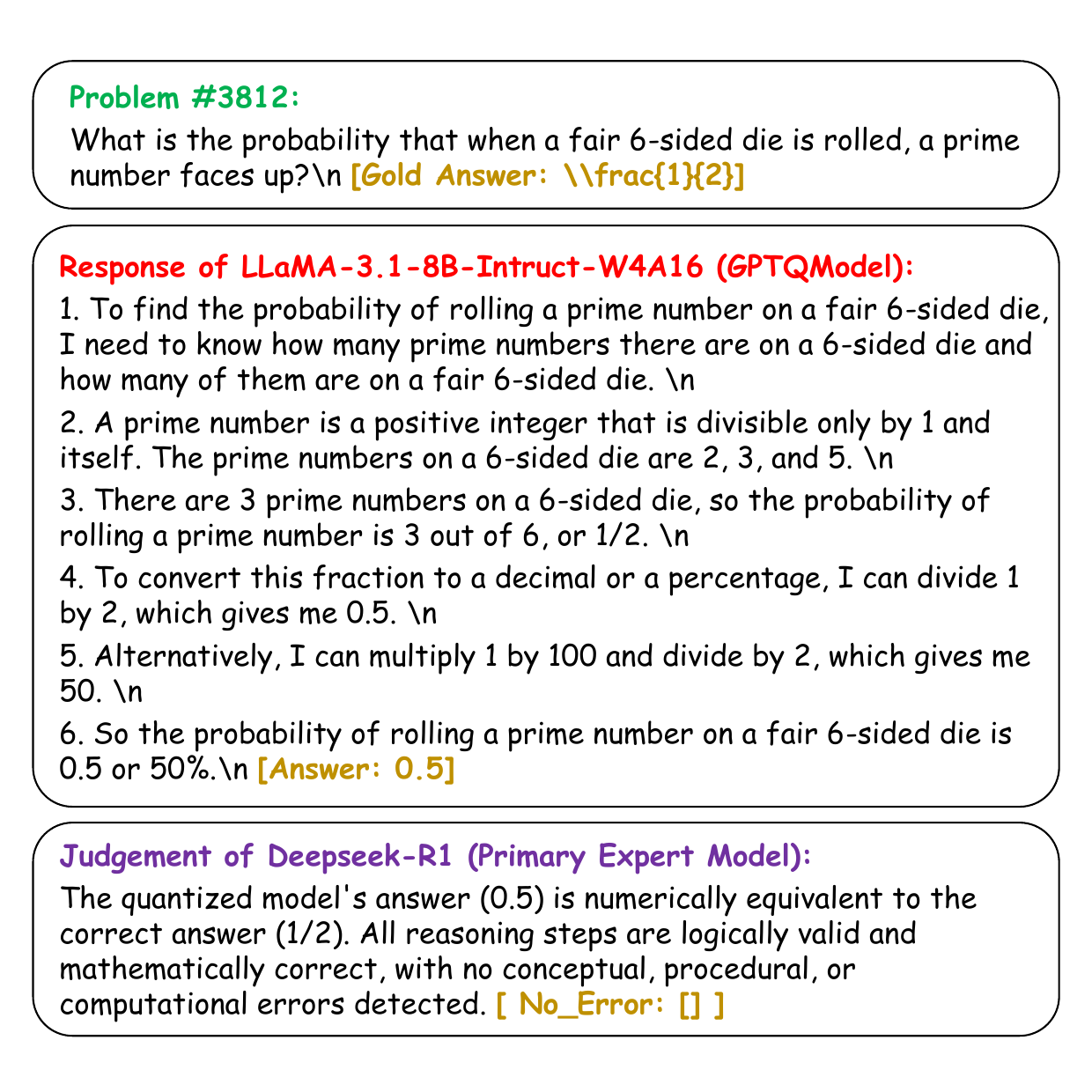}
    \caption{Problem \#3812: No Error.}
    \label{fig:no_error}
\end{figure}

\label{sec:appendix}


\end{document}